\documentclass[10pt,twocolumn,letterpaper]{article}

\usepackage{cvpr}              

\usepackage[utf8]{inputenc} 
\usepackage[T1]{fontenc}    
\usepackage{url}            
\usepackage{booktabs}       
\usepackage{amsfonts}       
\usepackage{nicefrac}       
\usepackage{microtype}      
\usepackage{xcolor}         

\usepackage{graphicx} 
\usepackage{caption}
\usepackage{wrapfig}
\usepackage{multirow}
\usepackage{booktabs}
\usepackage{arydshln}
\usepackage{pifont}

\usepackage{amsmath}
\usepackage{algorithm}
\usepackage{algpseudocode}

\usepackage{listings}
\usepackage{xcolor}

\usepackage{color}
\usepackage{colortbl}
\usepackage{tabularx} 
\usepackage{pifont}
\newcommand{\cmark}{\ding{51}} 
\newcommand{\xmark}{\ding{55}} 

\definecolor{mycolor_blue}{HTML}{E7EFFA}
\definecolor{mycolor_green}{HTML}{E6F8E0}
\definecolor{mycolor_gray}{HTML}{ECECEC}
\definecolor{pearDark}{HTML}{2980B9}
\definecolor{mycitecolor}{RGB}{0, 153, 51} 

\definecolor{citecolor}{HTML}{0071BC}
\definecolor{linkcolor}{HTML}{ED1C24}

\definecolor{cvprblue}{rgb}{0.21,0.49,0.74}
\usepackage[pagebackref,breaklinks,colorlinks,allcolors=cvprblue]{hyperref}

\def\paperID{} 
\def\confName{CVPR}
\def\confYear{2026}

\NewDocumentCommand{\inlineimage}{O{0.5} m}{%
  \raisebox{-0.25\baselineskip}{\includegraphics[height=#1\baselineskip]{#2}}\hspace{-1pt}
}

\newcommand{\icon}{\raisebox{0\height}{\inlineimage[1.]{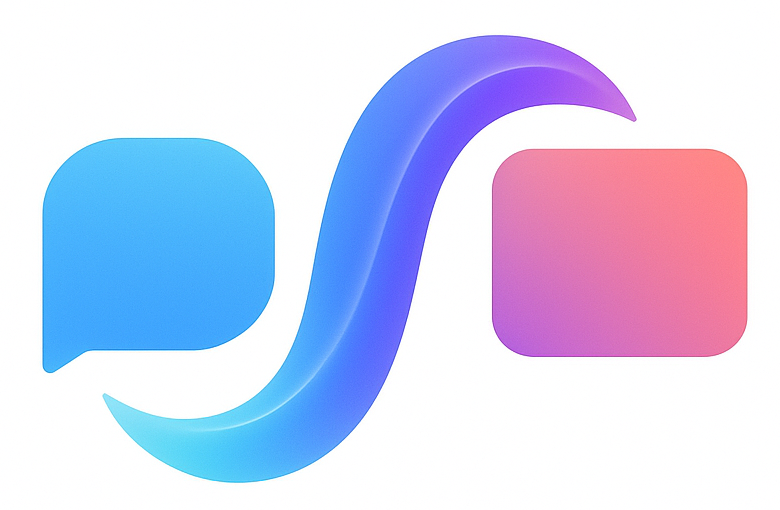}}}

\definecolor{echo_c1}{HTML}{5289fd}  
\definecolor{echo_c2}{HTML}{5869fd}
\definecolor{echo_c3}{HTML}{8a51f7}
\definecolor{echo_c4}{HTML}{c067d5}  
\definecolor{echo_c5}{HTML}{e479ac} 
\definecolor{echo_c6}{HTML}{fc8584}

\newcommand{\ours}{%
  \textcolor{echo_c1}{P}%
  \textcolor{echo_c2}{-}%
  \textcolor{echo_c3}{F}%
  \textcolor{echo_c4}{l}%
  \textcolor{echo_c5}{o}%
  \textcolor{echo_c6}{w}\xspace
}

\title{\icon \ours: Prompting Visual Effects Generation}
\author{Rui Zhao, Mike Zheng Shou\thanks{Corresponding Author.}\\ 
Show Lab, National University of Singapore 
}

\begin{document}
\maketitle

\begin{abstract}
Recent advancements in video generation models have significantly improved their ability to follow text prompts. However, the customization of dynamic visual effects, defined as temporally evolving and appearance-driven visual phenomena like object crushing or explosion, remains underexplored. Prior works on motion customization or control mainly focus on low-level motions of the subject or camera, which can be guided using explicit control signals such as motion trajectories. In contrast, dynamic visual effects involve higher-level semantics that are more naturally suited for control via text prompts. However, it is hard and time-consuming for humans to craft a single prompt that accurately specifies these effects, as they require complex temporal reasoning and iterative refinement over time. To address this challenge, we propose \ours, a novel training-free framework for customizing dynamic visual effects in video generation without modifying the underlying model. By leveraging the semantic and temporal reasoning capabilities of vision-language models, \ours performs test-time prompt optimization, refining prompts based on the discrepancy between the visual effects of the reference video and the generated output. Through iterative refinement, the prompts evolve to better induce the desired dynamic effect in novel scenes. Experiments demonstrate that \ours achieves high-fidelity and diverse visual effect customization and outperforms other models on both text-to-video and image-to-video generation tasks. Code is available at \href{https://github.com/showlab/P-Flow}{https://github.com/showlab/P-Flow}.

\end{abstract}

\section{Introduction}
\label{sec:intro}

\begin{figure}[!tb]
  \centering
\includegraphics[width=\linewidth]{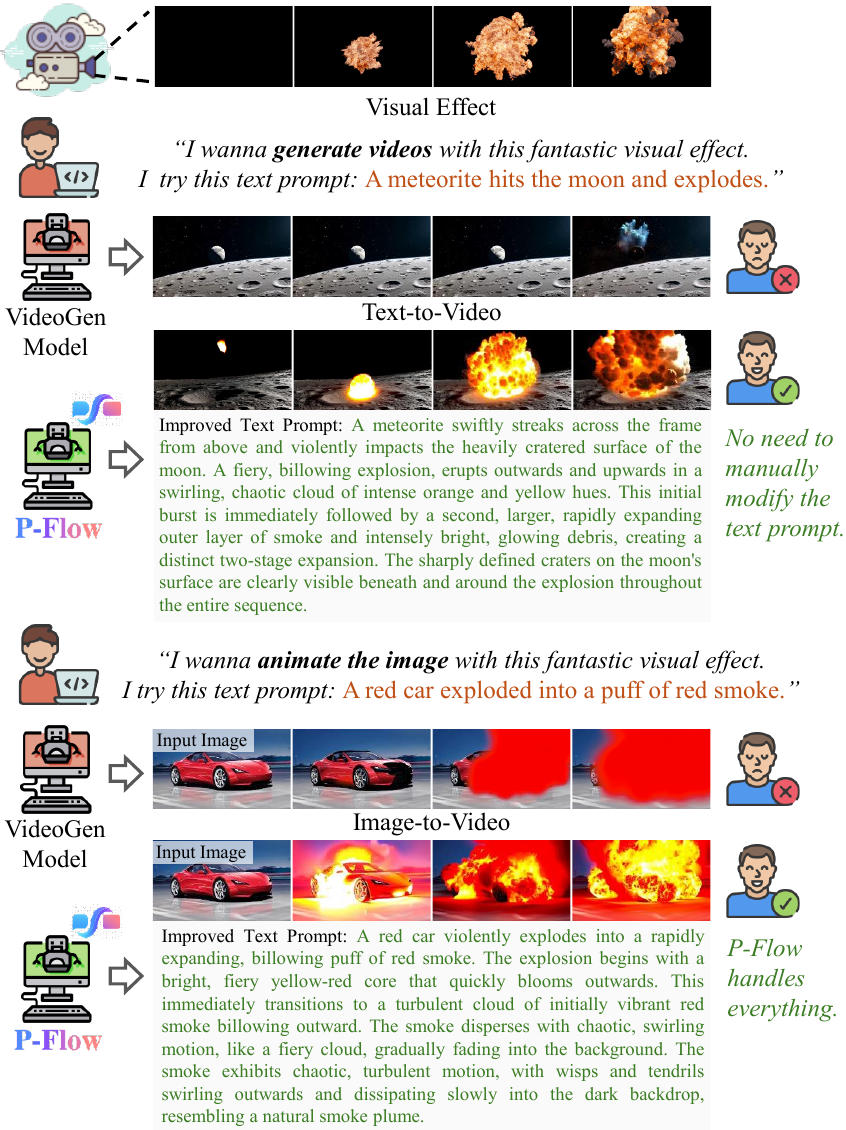}

\vspace{-0.3em}
\caption{
It is hard for humans to craft text prompts that precisely control video generation models to generate desired visual effects across diverse scenes, while \ours automatically refines prompts to achieve consistent and realistic visual effects.
}
\vspace{-0.5em}
\label{fig:teaser}
\end{figure}

Recent advancements in video generation models have significantly enhanced their ability to produce visually compelling content guided by text instructions~\cite{wang2025wan, kong2024hunyuanvideo, agarwal2025cosmos}. These models excel at generating videos that align with high-level semantic descriptions, enabling applications in creative storytelling, virtual environments, and visual design~\cite{bruce2024genie, feng2024matrix, guo2025mineworld, ren2025videoworld, yu2025gamefactory}. However, specifying nuanced, temporally evolving phenomena, such as \textit{dynamic visual effects} (e.g., object explosion, crushing), remains a challenge. Unlike low-level motion control~\cite{zhao2023motiondirector, wang2024motionctrl}, which can be guided by explicit trajectories, dynamic visual effects require higher-level semantic understanding and temporal coherence, making them difficult to capture with explicit conditions.

While such effects are naturally suited for control via text prompts due to their semantic richness, crafting prompts that accurately describe dynamic visual effects is inherently complex. Users must articulate both the semantic characteristics and temporal evolution of the effect, often requiring iterative refinement and complex temporal reasoning. For instance, applying a reference explosion effect to a new scene, such as a meteor crashing into the moon, requires preserving the dynamics and timing of the effect while adapting it to a completely different visual and semantic context, as shown in Fig.~\ref{fig:teaser}. Manual prompt engineering for such tasks is time-consuming and often yields suboptimal results.

Prior works on video customization have primarily focused on low-level motion control, such as guiding subject or camera motion using trajectories or spatial paths~\cite{wang2024motionctrl, hecameractrl}. While effective for explicit motion tasks, these methods are ill-suited for high-level semantic effects that lack clear motion trajectories. Alternative approaches that fine-tune video generation models for specific effects require extensive computational resources and lack generalizability across diverse effects~\cite{liu2025vfx}. 
In contrast, a training-free paradigm that leverages the powerful abilities of foundational generation models would offer a flexible and user-friendly solution for effect customization.

To address these challenges, we propose \ours, a novel training-free framework that customizes dynamic visual effects in video generation by treating text prompts as optimization variables. 
Rather than updating the generation model itself, \ours performs test-time prompt optimization, leveraging the semantic and temporal reasoning capabilities of vision-language models (VLMs) to iteratively refine prompts and bridge the gap between generated video and reference visual effects.
To make this optimization both effective and stable, we introduce two key strategies. 
First, we introduce a noise prior that emphasizes temporally salient dynamics in the reference effect to guide stable optimization, while incorporating stochastic noise to maintain diversity and exploration during prompt refinement.
Second, we incorporate a lightweight historical context mechanism that maintains past optimization trajectories, enabling more consistent and coherent refinement across iterations. Together, these designs ensure that prompts evolve meaningfully over time, achieving high-fidelity visual effects customization.

The experimental results validate the effectiveness and generality of \ours in enabling high-fidelity and diverse visual effect generation across both image-to-video and text-to-video generation settings. 
Without any model fine-tuning, \ours achieves state-of-the-art performance in key metrics such as FID-VID~\cite{unterthiner2018towards}, FVD~\cite{balaji2019conditional}, and Dynamic Degree~\cite{huang2024vbench}, and is strongly preferred in human evaluations. 
Compared to the training-based baseline constrained by fixed-length supervision and training dataset biases, our test-time optimization approach fully captures the temporal evolution of effects and better adapts to diverse scenes.
These findings demonstrate the potential of \ours as a plug-and-play solution for dynamic visual effect generation. 

Our code will be fully open-sourced. The main contributions are summarized as follows: (1) We propose \ours, a training-free framework that customizes dynamic visual effects in video generation by optimizing text prompts at test time. It supports both text-to-video and image-to-video generation. (2) We introduce a novel prompt optimization paradigm guided by VLM, enhanced with a noise prior to stabilize learning while preserving diversity, and a lightweight historical context mechanism to ensure optimization coherence. (3) Extensive experiments demonstrate the state-of-the-art performance of \ours across metrics and human evaluations.

\section{Related Works}

\begin{figure*}[!tb]
  \centering
\includegraphics[width=\linewidth]{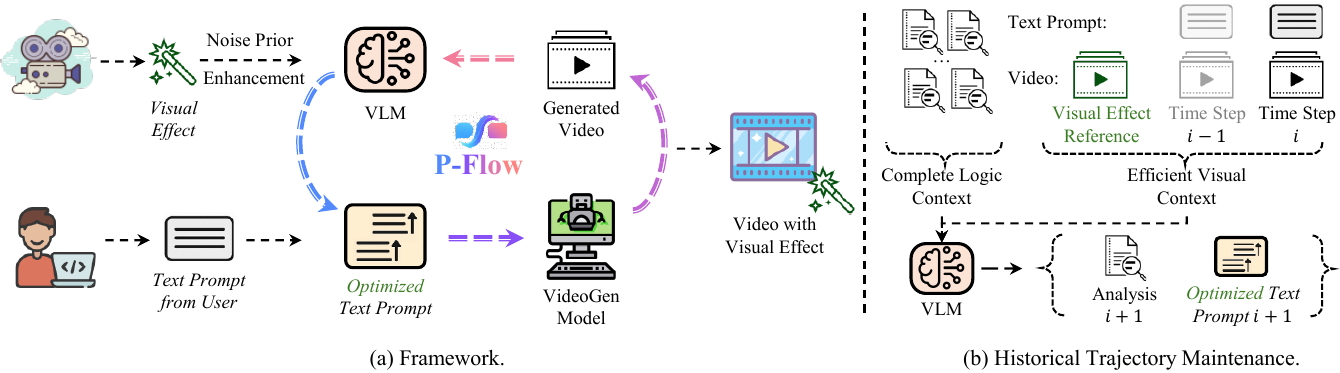}
\caption{Overview of the proposed \ours framework.
}
\vspace{-1em}
\label{fig:method}
\end{figure*}

\subsection{Video Generation Model}

Recent generation models demonstrate their powerful abilities in generating diverse and high- fidelity contents~\cite{ho2022imagen, singer2022make, he2022lvdm, VideoFusion, zhang2023show1, zhao2023zero, zhao2025doracycle}. Where video generation approaches are largely based on diffusion models~\cite{blattmann2023align, Wang2023LAVIEHV, vdm, videofactory, videoLDM, magicvideo, videogen, lavie, LVD, yu2023video, mei2023vidm}, which generate videos by denoising Gaussian noise through architectures such as 3D U-Net~\cite{unet} or transformer-based DiT~\cite{peebles2023scalable}. More recently, flow matching models~\cite{lipman2022flow, liu2022flow, jin2024pyramidal, ma2025step} have emerged as a scalable and efficient alternative, directly learning a velocity field to map noise to data without iterative denoising, and have shown superior quality on both realistic and diverse video generation tasks~\cite{kong2024hunyuanvideo, wang2025wan}. And a growing number of open-source video generation models ~\citep{zheng2024open, lin2024open, hacohen2024ltx, peng2025open, cogvideox, cogvideox} have recently been released, offering diverse architectures and capabilities for both text and image conditioned video generation.
The increasing fidelity and prompt-following ability of open-sourced SOTA models provide a promising foundation for prompt-based video generation and optimization.

\subsection{Motion Customization and Control}
Motion customization methods~\cite{zhao2023motiondirector} extend subject and style customization~\cite{ruiz2023dreambooth, kumari2023multi, gu2023mix, chen2023anydoor, wei2023elite, smith2023continual} to the temporal domain by enabling control over motion dynamics. DreamVideo~\cite{wei2024dreamvideo} and LAMP~\cite{wu2023lamp} learn motion patterns or adapters to customize both appearance and motion. Other methods~\cite{materzynska2023customizing, jeong2023vmc, wang2023motionctrl, ren2024customize, yang2024direct} further explore disentangled or reference-guided motion generation. In parallel, controllable video generation aims to ensure the generation results align with the given explicit control signals, such as depth maps, human pose, optical flows, etc. \cite{zhang2023adding, zhao2023uni, ma2023follow, Make-Your-Video, FollowYourPose,hecameractrl, zhang2024tora, bai2025recammaster, ren2025gen3c, wang2024motionctrl}.
These methods mainly address low-level motion using explicit priors or training-based control modules~\cite{wang2023videocomposer, chen2023control}. 

In contrast, dynamic visual effects involve higher-level semantics and remain underexplored. A concurrent work, VFX Creator~\cite{liu2025vfx}, adds control branches for visual effect generation but is limited to image-to-video generation and requires separate training for each different type of visual effect. Our method offers a flexible, training-free solution applicable to text-to-video and image-to-video models.

\subsection{Vision Language Models for Generation}

Recent advances in large language models~\cite{brown2020gpt3,openai2023gpt4,bai2023qwen,touvron2023llama,anil2023palm-2} have significantly enhanced the capabilities of vision-language models (VLMs)~\cite{bai2023qwen-vl,li2023blip-2,zhu2023minigpt-4,chen2023minigptv2,team2023gemini,liu2024visual}, enabling them to perform semantic and temporal reasoning over visual content. These models have been increasingly used to evaluate or guide  generation~\cite{yarom2024VNLI,lu2024llmscore,wiles2024Gecko,hu2023TIFA,cho2023DSG, manas2024improving,hao2023optimizing, mo2024dynamic, xiang2025promptsculptor, lee2024optimizing, cheng2025vpo, gao2025devil, nam2025optical, ji2025prompt, du2025vc4vg}, with Gecko~\cite{wiles2024Gecko} demonstrating their effectiveness in assessing fine-grained generation quality across diverse attributes. Some recent work, such as EvolveDirector~\cite{zhao2024evolvedirector} and VideoAlign~\cite{liu2025improving}, explored the use of VLM to train and optimize generation models to align human preferences. However, applying VLMs for test-time optimization in video generation remains largely unexplored. Our work leverages VLMs not only for evaluation but also as optimization tools to bridge the semantic gap between text prompts and complex visual effects.

\section{Method}
\label{sec:method}

\subsection{Problem Formulation}

Given a reference video \( V_{\text{ref}} \) showing a dynamic visual effect and an initial text prompt \( P_0 \) describing a novel scene or subject, our objective is to generate a video \( V_{\text{gen}} \) that exhibits the same visual effect as \( V_{\text{ref}} \) while adhering to the semantic content specified by \( P_0 \). 

In the image-to-video generation task, the generation is additionally conditioned on a source image \( I \) that provides detailed spatial structure or appearance of the scene. In this case, the generated video is given by \( V_{\text{gen}} = \mathcal{G}(P^*, I, \eta) \), where \( \mathcal{G} \) is a pre-trained video generation model and \( \eta \) denotes latent noise. For simplicity, and unless otherwise stated, we omit \( I \) in the formulation to unify notation across both text-to-video (T2V) and image-to-video (I2V) scenarios.

Formally, we aim to optimize a text prompt \( P^* \) such that the generated video \( V_{\text{gen}} = \mathcal{G}(P^*, \eta) \) minimizes the discrepancy \( \mathcal{D}(V_{\text{gen}}, V_{\text{ref}}) \) in terms of the semantic and temporal characteristics of the visual effect.

\subsection{Framework Overview}

The \ours framework operates in a training-free manner, optimizing the text prompt at test time without modifying the underlying video generation model. The method comprises three core components: (1) noise prior enhancement to initialize the latent noise for stable and diverse video sampling, (2) test-time prompt optimization using a VLM to iteratively refine the prompt, and (3) historical trajectory maintenance to guide the refinement decisions of VLM. The process is iterative, generating videos, evaluating their alignment with the reference effect, and refining the prompt until a maximum number of iterations is reached.

\subsection{Noise Prior Enhancement}

We found that the initial latent noise \( \eta \) used in video generation significantly influences optimization stability and output diversity. Completely random noise results in inconsistent visual effects across text prompt optimization iterations, hindering convergence, while fixed noise limits exploration, leading to suboptimal solutions. To address this, we propose a noise prior enhancement strategy that balances stability and exploration through flow matching inversion, temporal noise isolation, and noise blending.

First, we extract the latent noise corresponding to \(V_{\text{ref}}\) via flow matching inversion~\cite{rout2024semantic, kulikov2024flowedit, wang2024taming}.  In flow matching, the generative model defines a continuous-time ordinary differential equation (ODE)
\begin{equation}
\frac{\mathrm{d}x_t}{\mathrm{d}t}
\;=\;
v_\theta\bigl(x_t, t; P\bigr),
\end{equation}
which transports noise \(\eta\) at \(t=0\) to the data \(x_T\) at \(t=T\).  To invert this process, we integrate the same vector field backward in time starting from \(x_T = V_{\text{ref}}\) with its corresponding reference prompt \(P_{\text{ref}}\):
\begin{equation}
\eta_{\mathrm{inv}}
\;=\;
x_0
\;=\;
x_T
\;-\;
\int_{0}^{T}
v_\theta\bigl(x_t,\,t;\,P_{\mathrm{ref}}\bigr)\,\mathrm{d}t.
\end{equation}
By construction, this ensures
$\mathcal{G}\bigl(P_{\mathrm{ref}},\,\eta_{\mathrm{inv}}\bigr)
\;\approx\;
V_{\mathrm{ref}}$,
where \(\eta_{\mathrm{inv}}\) captures both the dynamic visual effect and appearance-specific attributes (e.g., textures or background elements) that are orthogonal to the visual effect itself.

To isolate the motion-related temporal components from the inverted noise \(\eta_{\mathrm{inv}} \in \mathbb{R}^{C \times F \times H \times W}\), where \( C \) is the number of latent channels, \( F \) is the number of frames, and \( H, W \) are spatial dimensions, we apply a two-stage SVD-based projection. 
First, we reshape \(\eta_{\mathrm{inv}}\) into a matrix \(\mathbf{N}_s \in \mathbb{R}^{(C \cdot F) \times (H \cdot W)}\) and compute its singular value decomposition:
\begin{equation}
\mathbf{N}_s = \mathbf{U}_s \mathbf{\Sigma}_s \mathbf{V}_s^\top.
\end{equation}
To suppress appearance-specific spatial variations, we adaptively determine the number of leading components \(k_s\) to remove by ensuring the retained energy satisfies
\begin{equation}
\frac{\sum_{i=k_s+1}^{r_s} \sigma_i^2}{\sum_{i=1}^{r_s} \sigma_i^2} \geq \rho_s,
\end{equation}
where \(r_s = \mathrm{rank}(\mathbf{N}_s)\). We set the top \(k_s\) singular values in \(\mathbf{\Sigma}_s\) to zero and reconstruct the spatially-filtered tensor as 
\begin{equation}
\eta_{\text{spatial}} = \mathrm{reshape}\left(\mathbf{U}_s \mathbf{\Sigma}'_s \mathbf{V}_s^\top, \, [C, F, H, W] \right).
\end{equation}

Next, \(\eta_{\text{spatial}}\) is reshaped along the temporal axis into \(\mathbf{N}_m \in \mathbb{R}^{(C \cdot H \cdot W) \times F}\) and do SVD project again, and we retain the top \(k_m\) components such that
\begin{equation}
\frac{\sum_{i=1}^{k_m} \sigma'_i{}^2}{\sum_{i=1}^{r_m} \sigma'_i{}^2} \geq \rho_m.
\end{equation}
The final projected noise \(\eta_{\text{temporal}} \in \mathbb{R}^{C \times F \times H \times W}\) preserves dominant motion information while suppressing static and appearance-dependent details.

Finally, to ensure exploratory diversity, we blend \( \eta_{\text{temporal}} \) with random noise \( \eta_{\text{new}} \sim \mathcal{N}(0, I) \):

\begin{equation}
\eta = \sqrt{\alpha} \cdot \eta_{\text{temporal}} + \sqrt{1 - \alpha} \cdot \eta_{\text{new}},
\end{equation}

where \( \alpha\) controls the influence of the motion-preserving noise. This blended noise \( \eta \) is used to sample the video \( V_{\text{gen}} = \mathcal{G}(P_i, \eta) \) at iteration \( i \). 

\subsection{Test-Time Prompt Optimization}

At each iteration \( i \), we generate a video \( V_{\text{gen}}^i \) using the current prompt \( P_i \) and the enhanced noise \( \eta \) as
\begin{equation}
    V_{\text{gen}}^i = \mathcal{G}(P_i, \eta),
\end{equation}
where $\mathcal{G}$ is the video generation model. 
To assess the alignment between the generated visual effects and those in the reference video \( V_{\text{ref}} \), we construct a composite video by vertically stacking \( V_{\text{ref}} \), the previously generated video (if available), and \( V_{\text{gen}}^i \). The composite video $V_\text{comb}$ is preprocessed to ensure consistent resolution and frame rate, enabling direct visual comparison across inputs.

A VLM is employed to analyze differences between \( V_{\text{gen}}^i \) and \( V_{\text{ref}} \), focusing on motion dynamics and visual effects, while explicitly ignoring variations in appearance or identity. Based on this analysis, the VLM performs prompt refinement to guide the next generation toward better reproducing the target visual effects:

\begin{equation}
P_{i+1} = \mathcal{M}(V_{\text{comb}}, P_i, \mathcal{H}; P_0)
\end{equation}

Here, \( \mathcal{M}(\cdot) \) denotes the VLM structured refinement function, which takes as input the reference and generated video pair $V_\text{comb}$, the current prompt \( P_i \), the historical trajectory of optimization, detailed in Sec.~\ref{sec:Historical}, and the original content constraints from \( P_0 \). The output is an updated prompt \( P_{i+1} \), where only effect-related descriptions are modified, preserving the original subject and environment.

The VLM is instructed to return a structured JSON object containing detailed analysis and the revised prompt \( P_{i+1} \). This iterative process enables fine-grained control over visual effect fidelity through prompt optimization. 
The full procedure is presented as pseudocode in the Appendix.

\begin{table*}[t]
\centering
\caption{Quantitative comparisons for both image-to-video and text-to-video generation settings. Note that VFX Creator supports only image-to-video generation.
}
\label{tab:quant_results}
\resizebox{1.0\textwidth}{!}{%
\begin{tabular}{lccccccccc}
\toprule
 
& \multicolumn{3}{c}{\textbf{Image-to-Video}} 
& \multicolumn{3}{c}{\textbf{Text-to-Video}}  
& \multicolumn{3}{c}{\textbf{Overall}} \\
\cmidrule(lr){2-4} \cmidrule(lr){5-7} \cmidrule(lr){8-10}
& FID-VID $\downarrow$ & FVD $\downarrow$ & Dyn. Degree $\uparrow$
& FID-VID $\downarrow$ & FVD $\downarrow$ & Dyn. Degree $\uparrow$ 
& FID-VID $\downarrow$ & FVD $\downarrow$ & Dyn. Degree $\uparrow$ \\
\midrule
Wan 2.1~\cite{wang2025wan}        & 34.62 & 994.70 & 0.33 & 42.32 & 1535.43 & 0.28 & 38.47 & 1265.07 & 0.31  \\
HunyuanVideo~\cite{kong2024hunyuanvideo}           & 36.53  & 1169.9 & 0.51 & 38.35  & 1362.35 & 0.34 & 37.44 & 1266.13 & 0.43 \\
VFX Creator \small{(\textit{Training-Based})}~\cite{liu2025vfx}      & 29.92  & \textbf{752.95}  & 0.63 & -- & -- & -- & -- & -- & -- \\
Wan 2.1 + HF \small{(\textit{Training-Free})}     & 33.12  & 989.42  & 0.54 & 42.21 & 1632.10 & 0.59 & 75.33 & 1310.76 & 0.57 \\
HunyuanVideo + HF \small{(\textit{Training-Free})}     & 33.85  & 1035.49  & 0.70 & 36.54 & 1266.79 &  0.62 & 35.20 & 1151.14 & 0.66 \\
\ours \small{(Ours, \textit{Training-Free})}   & \textbf{29.32}  &  784.51 & \textbf{0.94} & \textbf{32.93}  & \textbf{980.75} & \textbf{0.87} & \textbf{31.13} & \textbf{882.63}  & \textbf{0.91} \\
\bottomrule
\end{tabular}%
}
\end{table*}

\subsection{Historical Trajectory Maintenance}
\label{sec:Historical}

To enhance the reasoning and optimization capabilities of the VLM, we maintain a historical trajectory 
\begin{equation}
    \mathcal{H} = \{(V_i, P_i, A_i)\}_{i=0}^{i_{\max}-1}, 
\end{equation}
where \( V_i \), \( P_i \), and \( A_i \) denote the generated video, the corresponding prompt, and the VLM analysis at iteration \( i \). This trajectory provides context for prompt refinement, allowing the VLM to identify effective optimization patterns and avoid redundant changes. For example, if previous iterations have consistently increased the intensity of a desired visual effect, the VLM may favor similar refinements in subsequent steps.

However, storing the full sequence of previously generated videos introduces considerable computational overhead, especially as video inputs consume large amounts of visual tokens in the VLM. To address this, we adopt a memory-efficient strategy: only the reference video, the generated video from the previous iteration, and the current generated video are included in the visual input to the VLM. This selection maintains the most relevant temporal context while significantly reducing token length.

Meanwhile, to preserve long-term memory and optimization history, we retain all text prompts \( \{P_i\} \) and VLM analyses \( \{A_i\} \) across iterations in \( \mathcal{H} \). Since language tokens are much more compact than visual tokens, this design provides a good trade-off between efficiency and contextual richness, enabling the VLM to reason over past refinements while operating within practical computational constraints.

\subsection{Implementation Details}
\label{sec:Implementation Details}

We use the pre-trained Wan 2.1 14B video generation models~\cite{wang2025wan} for both text-to-video and image-to-video tasks, producing videos at the resolution of \(480 \times 832\) with 81 frames. For image-to-video generation, the aspect ratio is adaptively adjusted to be the same as the input image. 
Text prompt optimization is performed using the Gemini 1.5 Pro vision-language model. The blending weight is fixed to \( \alpha = 0.001 \), and the optimization process is run for \( i_{\max} = 10 \) iterations.  All experiments are conducted on an NVIDIA A100 GPU cluster. Video generation is performed with 8-GPU distributed inference, taking approximately 69 seconds per video and consuming around 40 GB of GPU memory per card. In each optimization iteration, besides the video generation, 1.2 seconds are used to construct the input for VLM, and 16.3 seconds are spent on prompt refinement via VLM inference. Structured instructions for VLM and more details are provided in the Appendix.

\section{Experiments}

We conduct comprehensive experiments to evaluate the effectiveness of \ours in customizing dynamic visual effects for video generation. 
The evaluation spans a diverse set of visual effects and includes both objective metrics and subjective human preference studies. We compare \ours with recent state-of-the-art methods through quantitative results in Sec.~\ref{sec:quantitative_results} and qualitative visualizations in Sec.~\ref{sec:Qualitative_Results}. In addition, we perform an ablation study in Sec.~\ref{sec:Ablation_Study} to analyze the contribution of key components. Experimental settings are detailed in Sec.~\ref{sec:exp_setup}.

\begin{figure*}[!tb]
  \centering
\includegraphics[width=\linewidth]{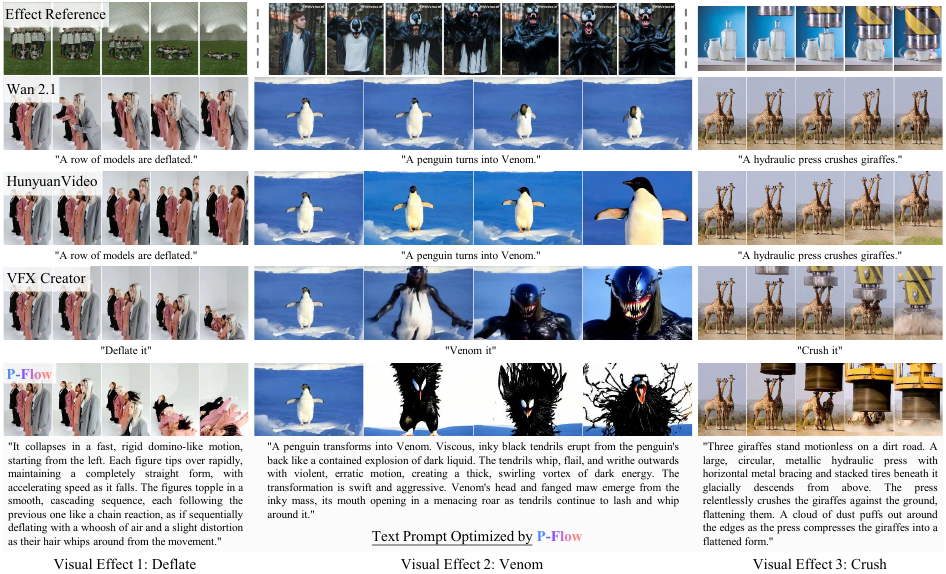}
\vspace{-1.5em}
\caption{Qualitative comparison on image-to-video generation with different visual effects. 
The prompts shown beneath each row represent the actual input to each model. As VFX Creator is optimized for short phrase inputs, such prompts are provided to ensure a fair and consistent evaluation.
}
\vspace{-1em}
\label{fig:qual_results}
\end{figure*}

\subsection{Experimental Setup}
\label{sec:exp_setup}

\textbf{Dataset}: The experiments are conducted on the Open-VFX dataset~\cite{liu2025vfx}. This benchmark comprises 675 high-quality videos sourced from commercial platforms, where each video lasts approximately 5 seconds at 24 fps. These videos span 15 diverse categories of dynamic visual effects, such as explode, deflate, and squish, offering rich visual diversity and temporal dynamics. Additionally, 245 reference images are provided for the image-to-video generation task, covering both single and multi-object scenes. We sample reference videos from its training set and test images from its test set.

\textbf{Metrics}: 
To assess the visual effect fidelity and dynamism of generated videos, we adopt three standard metrics following prior work~\cite{liu2025vfx}:
FID-VID~\cite{unterthiner2018towards}: Fréchet Inception Distance adapted for videos, measuring distributional similarity between generated and ground-truth videos. FVD~\cite{balaji2019conditional}: Fréchet Video Distance, which captures temporal coherence and realism based on a 3D video feature extractor. Dynamic Degree~\cite{huang2024vbench}: Quantifies the degree of motion or visual transformation across frames to reflect effect intensity and temporal variation.

In addition, we conduct a human evaluation using a pairwise comparison protocol, where 15 annotators are asked to choose the better video between two candidates in terms of visual effect fidelity. For each generation task, we sampled 100 samples with 15 different types of visual effects from each model.

\textbf{Baselines}: 
We compare \ours against the foundational state-of-the-art video generation models, Wan 2.1~\cite{wang2025wan} and HunyuanVideo~\cite{kong2024hunyuanvideo}, as well as a prior specialized model, VFX Creator~\cite{liu2025vfx}, which is specifically designed for visual effect learning. On the Open-VFX dataset, VFX Creator is trained with a separate LoRA version for each type of visual effect.
All baselines are used with their publicly released checkpoints and configurations.
We additionally include a human feedback (HF) mode for Wan 2.1 and HunyuanVideo, where the text prompt is manually revised once, based on the generated results, to improve the visual alignment with the given visual effect references.

\subsection{Quantitative Results}
\label{sec:quantitative_results}

As shown in Table~\ref{tab:quant_results}, our proposed method \ours achieves superior or highly competitive performance across all metrics in both image-to-video and text-to-video generation tasks. \ours is built upon Wan 2.1 in our experiments.

\begin{table}[t]
\centering
\caption{Human evaluation results (\%) comparing \ours against baseline models in Image-to-Video (I2V) and Text-to-Video (T2V) generation. Note: Model order was randomized during evaluation.}
\label{tab:human_eval}

\resizebox{1.0\linewidth}{!}{
\begin{tabular}{lcl}
\toprule
\textbf{Model 1} & \textbf{Preference (Model 1 V.S. Model 2)} & \textbf{Model 2} \\
\midrule
\ours-I2V & \textbf{80\%} V.S. 20\% & Wan 2.1-I2V \\
\ours-I2V & \textbf{84\%} V.S. 16\% & HunyuanVideo-I2V \\
\ours-I2V & \textbf{58\%} V.S. 42\% & VFX Creator \\
\ours-T2V & \textbf{75\%} V.S. 25\% & Wan 2.1-T2V \\
\ours-T2V & \textbf{81\%} V.S. 19\% & HunyuanVideo-T2V \\
\bottomrule
\end{tabular}

}
\end{table}

Specifically, \ours outperforms strong foundational video generation models such as Wan 2.1 and HunyuanVideo across all three metrics in both generation settings. Notably, our method achieves this without any fine-tuning or modification of the foundational model parameters, demonstrating the effectiveness of our test-time optimization framework. This validates our design philosophy of treating the video generator as a black box while still enabling high-quality visual effect generation through adaptive, input-specific optimization.

Compared to the training-based method VFX Creator, which is trained on the Open-VFX dataset and involves dedicated architectural designs, our method achieves comparable results in FID-VID and FVD, while significantly outperforming it in Dynamic Degree. This highlights the strength of our method in generating videos with more salient and temporally coherent motion, which is essential for visual effects generation.

Moreover, it is worth noting that VFX Creator does not support text-to-video generation, and its trained LoRA weights are tightly coupled with specific architectures. In contrast, \ours is training-free, modular, and model-agnostic, supporting both image-to-video and text-to-video tasks. Achieving such generalization and performance without any training overhead underscores the practicality and robustness of our framework.

A pairwise human preference study is conducted to compare the visual effect fidelity of \ours with others. As shown in Table~\ref{tab:human_eval}, the results demonstrate that \ours consistently outperforms existing models in both settings, reflecting its superiority in visual effect generation.

\subsection{Qualitative Results}

\label{sec:Qualitative_Results}

\renewcommand{\tabcolsep}{1.5pt}
\def\swfive{0.30\linewidth}

\begin{figure*}[!tb]
  \centering
\includegraphics[width=\linewidth]{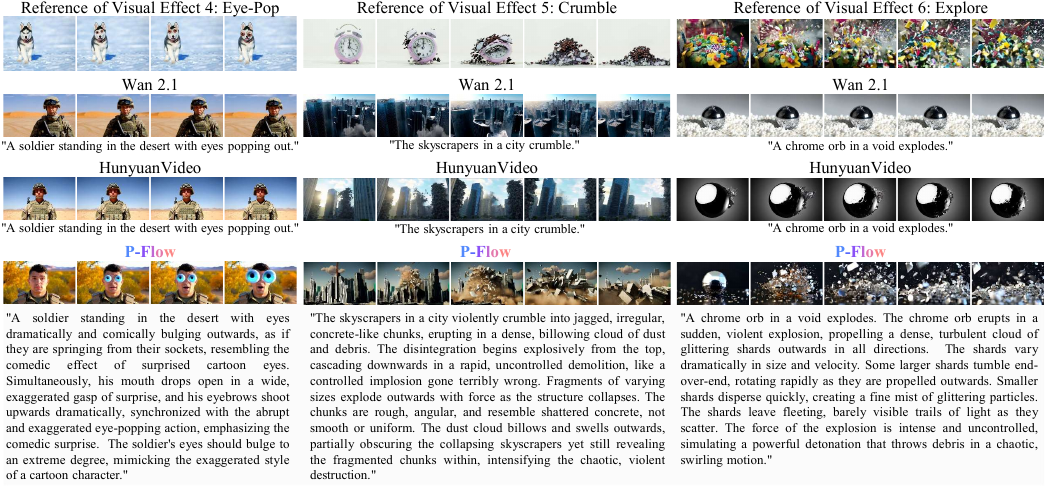}
\caption{Qualitative comparison on image-to-video generation with different visual effects. 
The prompts shown beneath each row represent the actual input to each model.
}
\label{fig:results_3}
\end{figure*}

\begin{figure*}[!tb]
  \centering
\includegraphics[width=\linewidth]{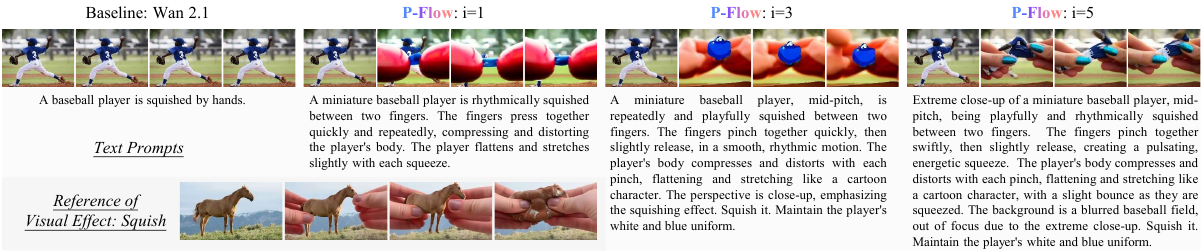}
\vspace{-1.5em}
\caption{Optimization trajectory of~\ours. Starting from a simple base prompt, \ours iteratively refines the text prompt based on the visual feedback (we showcase the iteration 1 → 3 → 5), leading to progressively more accurate alignment between the generated video and the target “squish” visual effect.
}
\label{fig:results_2}
\end{figure*}

As shown in Fig.~\ref{fig:qual_results} and Fig.~\ref{fig:results_3}, our proposed \ours demonstrates clear advantages in generating high-quality and controllable visual effects. 
It is worth mentioning that, for the \ours, there are no constraints on the resolution or length of the reference video. This greatly reduces the barrier for users to adopt our method, allowing them to freely choose reference clips of any duration or resolution.

The pre-trained strong foundational models, Wan 2.1 and HunyuanVideo, fail to produce the desired effects using plain text prompts, which highlights the insufficiency of generic prompts in steering these models toward specific visual goals.

In comparison, the training-based model, VFX Creator, exhibits relatively stronger ability in capturing visual effects. Nevertheless, it also suffers from inherent limitations imposed by its fixed-length training regime. For example, in the \textit{Visual Effect 1: Deflation}, the synthesized sequence terminates before the visual transformation completes. This truncation is attributed to that all training samples are forcibly trimmed to a fixed length, which may lead to cutting out some parts of the visual effects. \ours, in contrast, imposes no such constraint. The full reference video can be encoded by the VLM, allowing the dynamic evolution of the effect to be fully captured and reflected in the optimized prompt, thereby avoiding truncation-related failures. 

In addition, the training-based method may also encode dataset-specific biases. For example, in \textit{Visual Effect 2: Venom}, the second frame generated by VFX Creator includes a humanoid body structure, likely due to bias in the training data toward human-centric subjects. These artifacts reveal the limited generalization capacity of training-based models under distribution shifts. Our method, by optimizing the prompt at inference time based on the input image and reference video, naturally avoids such artifacts, accurately preserving subject-specific attributes from the input image while incorporating the desired visual effect from references.
The results in Fig.~\ref{fig:results_3} further demonstrate the superiority of our method on the text-to-video generation task.

\begin{table*}[t]
\centering
\caption{Ablation study of \ours on both image-to-video and text-to-video generation.}
\label{tab:ablation_full}
\resizebox{1.0\textwidth}{!}{%
\begin{tabular}{ccccccccccccc}
\toprule

\multicolumn{4}{c}{\textbf{Modules}} & \multicolumn{3}{c}{\textbf{Image-to-Video} }
& \multicolumn{3}{c}{\textbf{Text-to-Video}}  
& \multicolumn{3}{c}{\textbf{Overall}} \\

\cmidrule(lr){4-7} \cmidrule(lr){8-10} \cmidrule(lr){11-13}
Noise-Enhance & Logic-Context & Visual-Context (i-1) &  & FID-VID $\downarrow$ & FVD $\downarrow$ & Dyn. Degree $\uparrow$ & FID-VID $\downarrow$ & FVD $\downarrow$ & Dyn. Degree $\uparrow$ & FID-VID $\downarrow$ & FVD $\downarrow$ & Dyn. Degree $\uparrow$ \\
\midrule
\xmark & \xmark & \xmark & & 33.42 & 1089.23 & 0.64 & 39.85 & 1321.70 & 0.61 &36.64 & 1205.47 & 0.63\\
\cmark & \xmark & \xmark & & 32.80 & 961.78 & 0.69 & 36.74 & 1182.42 & 0.66 &34.77 &1072.10 & 0.68 \\
\cmark & \xmark & \cmark & & 30.45 & 861.52 & 0.83 & 34.05 & 1044.67 & 0.78  & 32.25 &953.10 &0.81 \\
\cmark & \cmark & \cmark & & \textbf{29.32} & \textbf{784.51} & \textbf{0.94} & \textbf{32.93} & \textbf{980.75} & \textbf{0.87} &\textbf{31.13} & \textbf{882.63} & \textbf{0.91}\\
\bottomrule
\end{tabular}%
}

\vspace{-1em}
\end{table*}

\textbf{Optimization Trajectory.} 
We visualize the prompt optimization trajectory of \ours in Fig.~\ref{fig:results_2}. Given a reference video containing the desired visual effect, \ours gradually optimizes the text prompt to guide the generation towards similar dynamics in a novel scene.

\subsection{Ablation Study}

\label{sec:Ablation_Study}

We conduct the ablation study to investigate the effectiveness of each component in our framework, including the Noise-Enhance module, the Visual-Context (i-1), and the Logic-Context modules. Results are summarized in Table~\ref{tab:ablation_full} under both image-to-video and text-to-video settings. Specifically, Visual-Context (i-1) refers to incorporating the previously generated video frame at time step i-1 as visual context for the current generation.

It is shown that even without incorporating any of the three ablation components, the performance of \ours already surpasses the strong foundational model, Wan 2.1~\cite{wang2025wan}, in terms of Dynamic Degree. This demonstrates that text prompt optimization alone, without any tuning or additional temporal modules, can significantly enhance the temporal dynamics.

As shown in Table~\ref{tab:ablation_full}, each module contributes incrementally to performance. Adding the Noise-Enhance component leads to improvements because it can stabilize the optimization of our framework. Introducing short-term context through the Visual-Context module brings further gains by offering visual insights for VLM to better analyze the influence of the text prompt and further optimize it. Finally, we incorporate the Logic-Context module, which provides long-range semantic analysis context derived from the entire optimization trajectory. This allows the prompt to maintain high-level coherence and effect progression over time. 
Notably, by decoupling long-term logic context from short-term visual context, our method avoids the computational overhead of processing long visual sequences, while still benefiting from both temporal scales.

\subsection{Hyperparameter Analysis}

Our noise prior enhancement strategy involves hyperparameters that control the trade-off between optimization stability and generation diversity. We conduct a parameter study on mage-to-video generation to analyze their effects, and summarize the results as follows.

\textbf{Energy Thresholds for SVD Projection.} 
We introduce two energy thresholds, \( \rho_s \) and \( \rho_m \), to determine the number of principal components retained or suppressed during the two-stage SVD-based projection:
\begin{itemize}
  \item Spatial energy threshold (\( \rho_s \)): Controls the suppression of appearance-related spatial details, e.g., textures, tone, and background patterns. 
  \item Temporal energy threshold (\( \rho_m \)): Determines the amount of motion-relevant temporal variation to retain.

\end{itemize}

When setting \( \rho_s = 0 \), i.e., without spatial suppression, the model retains unwanted appearance priors from the reference video, resulting in degraded visual quality, i.e. FID-VID: 33.25 and FVD: 1052.80, as shown in Table~\ref{tab:ablation_noise_image2video}. On the other hand, we empirically observed that setting \( \rho_s \) too high (\( > 0.5 \)) overly suppresses useful priors, leading to diminished impact of the enhanced noise. A moderate value $\rho_s = 0.1$ achieves the best balance.

For the temporal energy threshold, we set $\rho_m = 0.9$ to retain most of the motion-relevant information. As shown in Table~\ref{tab:ablation_noise_image2video}, these settings help preserve temporal dynamics and achieve a good dynamic score as 0.94.

\textbf{Blending Coefficient \( \alpha \).} 
We further study the impact of the blending coefficient \( \alpha \in [0,1] \), which controls the mixture of preserved temporal noise and fresh random noise. 

As shown in Table~\ref{tab:ablation_noise_image2video}, using only random noise \( \alpha = 0 \) achieves limited performance because of the unstable optimization process. 
A suitable coefficient \( \alpha = 0.001 \) leads to significant improvement across all metrics, including FVD and motion dynamics, by preserving key information of motion dynamics while introducing sufficient randomness. Slightly increasing \( \alpha \) to 0.01 further improves FID-VID but reduces dynamic scores, reflecting a trade-off between fidelity and motion dynamics. We finally set \( \alpha \) to 0.001 to achieve better dynamical generation.

\begin{table}[t]
\centering
\caption{Analysis of noise prior enhancement components.}
\vspace{-0.5em}
\label{tab:ablation_noise_image2video}
\resizebox{0.46\textwidth}{!}{%
\begin{tabular}{lccc}
\toprule
\textbf{Setting} & FID-VID $\downarrow$ & FVD $\downarrow$ & Dyn. Degree $\uparrow$ \\
\midrule
w/o SVD Projection (\( \rho_s=0.0, \rho_m=1.0 \)) & 33.25 & 1052.80 & 0.58 \\
Random Noise Only (\( \alpha = 0.0 \)) & 32.74 & 923.67 & 0.73 \\
Enhanced Noise (\( \alpha = 0.001, \rho_s=0.1, \rho_m=0.9 \)) & 29.32 & \textbf{784.51} & \textbf{0.94} \\
Enhanced Noise (\( \alpha = 0.01, \rho_s=0.1, \rho_m=0.9 \)) & \textbf{29.21} & 803.35 & 0.88 \\
\bottomrule
\end{tabular}%
}

\vspace{-0.8em}
\end{table}

\section{Conclusion}
We present \ours, a training-free framework for customizing dynamic visual effects in video generation through test-time prompt optimization. By leveraging noise prior enhancement and historical trajectory, \ours enables stable and coherent effect transfer without model fine-tuning. Extensive experiments demonstrate its strong performance and generality, highlighting \ours as a practical framework for generating high-fidelity visual effects at test-time.


{
    \small
    \bibliographystyle{ieeenat_fullname}
    \bibliography{main}
}

\definecolor{codegray}{RGB}{245,245,245}   
\definecolor{codeblue}{RGB}{30,70,160}     
\definecolor{codegreen}{RGB}{0,128,0}     
\definecolor{codered}{RGB}{180,0,0}       
\definecolor{codeblack}{RGB}{0,0,0}  

\lstdefinestyle{promptblock}{
  language={},                     
  basicstyle=\small\rmfamily\color{codegreen},  
  backgroundcolor=\color{codegray},
  frame=single,
  frameround=tttt,
  rulecolor=\color{gray},
  breaklines=true,
  columns=fullflexible,
  showstringspaces=false,
  captionpos=b,
}

\clearpage
\setcounter{page}{1}
\maketitlesupplementary

\appendix

\begin{algorithm*}
\caption{\ours Framework}
\label{alg:alg1}
\begin{algorithmic}[1]

\Require Reference video \( V_{\text{ref}} \), initial prompt \( P_0 \), optional input image \( I \), video diffusion model \( \mathcal{G} \), VLM \( \mathcal{M} \), max iterations \( i_{\max} \), blending weight \( \alpha \)
\State Initialize historical trajectory \( \mathcal{H} \gets \emptyset \), iteration index \( i \gets 0 \)
\State Compute inversion noise \( \eta_{\text{inv}} \gets \text{FlowMatchingInversion}(V_{\text{ref}}, P_0, I, \mathcal{G}) \)
\State Compute temporal noise \( \eta_{\text{temporal}} \gets \text{ProjectNoiseTemporally}(\eta_{\text{inv}}) \)
\State Set current prompt \( P_i \gets P_0 \)
\For{\( i < i_{\max} \)}
    \State Sample random noise \( \eta_{\text{new}} \sim \mathcal{N}(0, I) \)
    \State Blend noise \( \eta \gets \sqrt{\alpha} \cdot \eta_{\text{temporal}} + \sqrt{1 - \alpha} \cdot \eta_{\text{new}} \)
    \State Generate video \( V_i \gets \mathcal{G}(P_i, I, \eta) \)
    \If{\( i = 0 \)}
        \State Combine videos \( V_{\text{comb}} \gets \text{CombineVideos}([V_{\text{ref}}, V_i]) \)
    \Else
        \State Combine videos \( V_{\text{comb}} \gets \text{CombineVideos}([V_{\text{ref}}, V_{i-1}, V_i]) \)
    \EndIf
    \State Analyze and refine prompt: \( (A_i, P_{i+1}) \gets \mathcal{M}(V_{\text{comb}}, P_i, \mathcal{H}) \)
    \State Update history: \( \mathcal{H} \gets \text{UpdateHistory}(\mathcal{H}, P_i, A_i, V_i) \)

    \State \( i \gets i + 1 \)
\EndFor

\end{algorithmic}

\Return Optimized prompt \( P_i \), generated video \( V_i \), trajectory \( \mathcal{H} \)
\end{algorithm*}

\section{Implementation Details}

\subsection{Prompt Optimization Procedure}
\label{supsec:procedure}

Algorithm~\ref{alg:alg1} summarizes the full test-time optimization loop used in \ours. 
Given a reference video $V_{\text{ref}}$, an initial prompt $P_0$, and an optional input image $I$, we first initialize the historical trajectory $\mathcal{H}$ and the iteration index $i$. 
We then compute an inversion noise code $\eta_{\text{inv}}$ by calling 
\texttt{FlowMatchingInversion} on $(V_{\text{ref}}, P_0, I, \mathcal{G})$, and obtain a motion-preserving temporal prior $\eta_{\text{temporal}}$ with 
\texttt{ProjectNoiseTemporally}$(\eta_{\text{inv}})$. 
These two steps implement the noise prior enhancement described in the method section of the main paper.

Starting from $P_i = P_0$, the algorithm performs an iterative refinement over $i = 0, \dots, i_{\max}-1$. 
At each iteration, we first sample a fresh Gaussian noise $\eta_{\text{new}} \sim \mathcal{N}(0, I)$ and blend it with the temporal prior to form the actual sampling noise
\[
\eta \;=\; \sqrt{\alpha}\,\eta_{\text{temporal}} + \sqrt{1-\alpha}\,\eta_{\text{new}},
\]
where $\alpha$ controls the trade-off between stability, reusing the temporal prior, and diversity and exploratory, introducing new randomness. 
Using this blended noise, the video diffusion model $\mathcal{G}$ generates a video 
$V_i = \mathcal{G}(P_i, I, \eta)$ conditioned on the current prompt and, when applicable, the input image.

To provide the VLM with a direct, visual comparison between the reference effect and the current generations, we construct a combined video $V_{\text{comb}}$ by concatenating multiple clips. 
In the first iteration, $V_{\text{comb}}$ contains only $V_{\text{ref}}$ and $V_0$; 
in subsequent iterations, it contains $V_{\text{ref}}$, the previous generation $V_{i-1}$, and the current one $V_i$. 
This design allows the VLM to assess both the absolute discrepancy with the reference and the incremental change across iterations.

The vision-language model $\mathcal{M}$ then takes $(V_{\text{comb}}, P_i, \mathcal{H})$ as input and 
returns a diagnostic analysis $A_i$ together with an updated prompt $P_{i+1}$. 
Here, $\mathcal{H}$ denotes the historical trajectory that stores past prompts, analyses, 
and generated videos, as described in the method section of the main paper.
We update $\mathcal{H}$ via \texttt{UpdateHistory} to include $(P_i, A_i, V_i)$, and proceed to the next 
iteration. 
After $i_{\max}$ iterations, the procedure outputs the final optimized prompt, the last generated video, 
and the complete trajectory $\mathcal{H}$ as summarized in Algorithm~\ref{alg:alg1}.

\subsection{Structured instruction for VLM}

The instruction provided to the VLM are detailed in Listing~\ref{lst:instruction}. It directs the VLM to analyze a combined video containing up to three segments (reference, last generated, and newly generated), compare their visual effects and motion dynamics, and refine the prompt to minimize the misalignments while preserving the subject and environment. The instruction operates by iteratively updating the \texttt{<current\_prompt>} with the refined text prompt based on the VLM analysis, leveraging a memory of past iterations \texttt{<memory\_to\_replace>} to track refinement effectiveness. 

The placeholders, such as \texttt{<current\_prompt>} and \texttt{<memory\_to\_replace>}, are dynamic variables, iteratively updated by the \ours. While the placeholders \texttt{<subject>} and \texttt{<environment>} are fixed and automatically extracted from the initial text prompt, and the \texttt{<desired\_visual\_effect>} is given by user input. The instruction mandates the VLM to output a structured JSON content, containing the analysis and refined prompt, enabling automated parsing and integration into the iterative pipeline.

\begin{figure*}[t]
\centering
\begin{minipage}{\textwidth}
\begin{lstlisting}[style=promptblock, caption={Instructions for VLM}, label={lst:instruction}]
Instruction = """ 
Your task is to optimize a text prompt for the video generation model to match the reference video's dynamic visual effect "<desired_visual_effect>".

Input: Combined video with up to three segments:
- "A" (top): Reference video.
- "B" (middle, if present): Last generated video. Corresponding text prompt: "<last_text_prompt>".
- "C" (bottom): New generated video. Corresponding text prompt: "<current_text_prompt>".

Steps:
1. **Analyze**:
   - "A": Describe visual effects (focusing on "<desired_visual_effect>" related dynamics), followed by related motion dynamics (speed, direction, pattern) and transitions (timing, rhythm).
   - "B" (if present): Summarize visual effects, motion dynamics, and transitions.
   - "C": Summarize visual effects, motion dynamics, and transitions.
   
2. **Compare**:
   - Compare "C" (and "B", if present) to "A" for differences in visual effects, motion dynamics, and transitions.
   - For "B", identify prompt terms causing misalignments in visual effects or motion dynamics.
   - Evaluate how the prompt changes from "B" to "C" affects the visual effects alignment with "A".

3. **Refine Prompt**:
   - Keep "<subject>" and "<environment>" unchanged.
   - Refine the text prompt "<current_prompt>" to match "A"'s visual effects "   <desired_visual_effect>", and related motion dynamics and transitions better, and fix its errors.
   - Avoid instructional language and problematic terms.

4. **Output**:
   - JSON:
     - "analysis":
       - "reference_description": "A"'s visual effects,  motion dynamics, and transitions.
       - "last_generated_description" (if "B" exists): "B"'s visual effects, motion dynamics, and transitions.
       - "new_generated_description": "C"'s visual effects, motion dynamics, and transitions.
       - "comparison": Summary of differences of "C" and "A" in visual effects, motion dynamics, and transitions, including errors in "B"'s prompt and their impact..
     - "refined_prompt": Optimized prompt for "C" to minimize the misalignment with "A"'s visual effects.


Guidelines:
- Use "<memory_to_replace>" to track the history of prompt refinements and their effectiveness.
- Prioritize "<desired_visual_effect>" and visual effects, then motion dynamics and transitions.
- Do not include non-visual effect details from "A" (e.g., specific colors or other appearance-related elements unless part of "<desired_visual_effect>").

Previous history: <memory_to_replace>
Subject: <subject>
Environment: <environment>
Desired Visual Effect: <desired_visual_effect>
Current prompt: <current_prompt>
"""

\end{lstlisting}
\end{minipage}
\end{figure*}

\section{Limitations and Future Works}

Despite the promising visual effect customization performance, our current framework still has limitations in terms of optimization efficiency. 
First, the number of optimization iterations is fixed across all cases, which may lead to suboptimal efficiency. In practice, we observe that some prompts can achieve satisfactory visual effects within a few iterations, while more challenging cases may require extended refinement. However, without an adaptive stopping mechanism, the optimization process may either run longer than needed or stop before achieving optimal results. In future work, we plan to introduce an auxiliary VLM as an evaluator to dynamically assess the alignment between the generated visual effect and the target one, thereby enabling adaptive stopping when sufficient quality is achieved.

Second, the current framework relies on full video generation through multiple flow-matching steps before evaluating the alignment with the desired visual effect. Combined with the iterative prompt optimization loop, this results in a relatively time-consuming process. Empirically, we find that the primary visual effects often emerge in the early part of the generation time steps. This motivates a future direction to perform evaluation and prompt refinement at intermediate generation time steps, potentially reducing time cost and improving overall efficiency.

\section{Potential Broader Implications}
\label{sec:broader-impacts}

We present a prompt optimization framework that enables visual effect customization in video generation. By improving the controllability of video outputs through natural language, our method lowers the barrier for users to generate videos with desired visual effects. This could benefit creative industries such as animation, marketing, and virtual content creation, while also advancing research in the customization of video generation.

However, as with all generative models, our framework inherits potential risks, including the amplification of societal biases and the possibility of misuse, such as generating misleading or harmful content. To mitigate these risks, we will include explicit terms of use in the user agreement, warning against the generation of violent, obscene, or deceptive content. These terms are intended to discourage unethical usage and clarify user responsibility when interacting with the system.

Besides, our framework builds upon pre-trained video generation models and vision-language models that have integrated safety checkers. These built-in mechanisms help detect and filter out undesirable outputs during generation.

\section{More Results}
\begin{figure*}[!tb]
  \centering
\includegraphics[width=\linewidth]{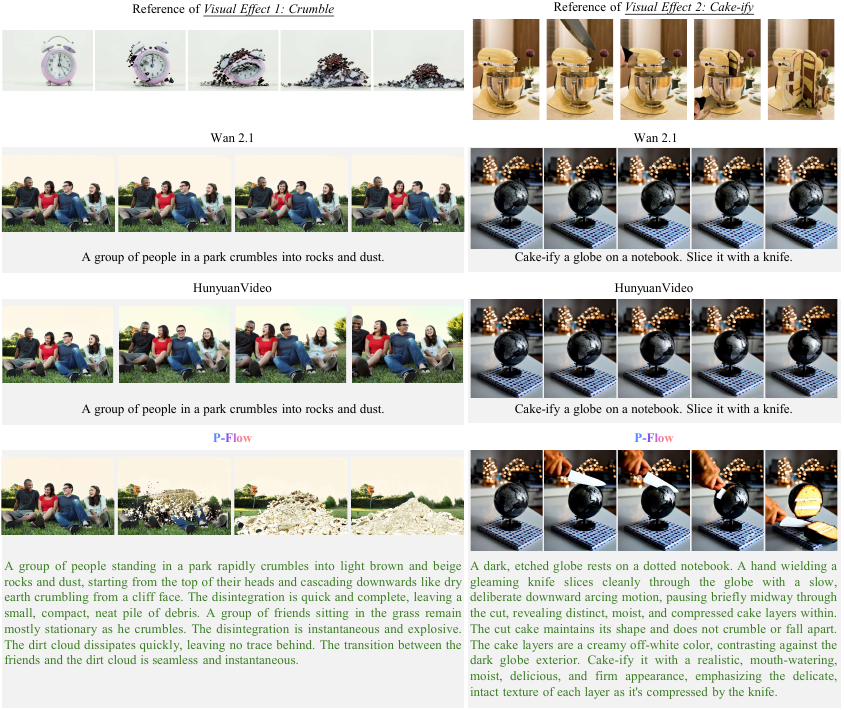}
\vspace{-1.5em}
\caption{Image-to-Video Generation Results.
}
\vspace{-1em}
\label{fig:appendix_results_i2v}
\end{figure*}

We provided more image-to-video generation results in  Fig.~\ref{fig:appendix_results_i2v} and text-to-video generation results in Fig.~\ref{fig:appendix_results_t2v}. The video version of the results presented in the appendix and main paper can be found in the zip file attached within the supplementary material.

In Fig.~\ref{fig:appendix_results_i2v}, we present image-to-video generation results on two challenging visual effects: \textit{Crumble} and \textit{Cake-ify}. 
Compared with Wan~2.1 and HunyuanVideo, both of which tend to either preserve the input appearance with minimal effect expression or generate inappropriate transformations, \ours\ achieves substantially more faithful and temporally consistent effect reproduction. 
Leveraging the refined prompts generated during optimization, our method is able to 
perceive visual cues from the input image while inducing effect behaviors that closely match the reference dynamics, for instance, controlled disintegration patterns in \textit{Crumble} or revealing the internal structure of an object in \textit{Cake-ify}. 
These results demonstrate that \ours\ maintains strong visual coherence with the source image while enabling expressive and high-fidelity dynamic visual effect customization.

In Fig.~\ref{fig:appendix_results_t2v}, we compare \ours\ with Wan~2.1 and 
HunyuanVideo on two dynamic visual effects: \textit{Levitate} and 
\textit{Inflate}. For each effect, we show the reference video, baseline generations, and the result generated by our method accompanied by the refined prompt. 
As illustrated, baseline models often generate weakly expressed motions that loosely resemble the intended visual dynamics, whereas \ours\ 
successfully induces high-fidelity, temporally coherent visual effects that more closely match the reference progression. 
The refined prompts generated by our method capture richer temporal and effect-related semantics, which enable the video generation model to reproduce more high-fidelity and expressive visual effect behaviors in novel scenes.

\begin{figure*}[!tb]
  \centering
\includegraphics[width=\linewidth]{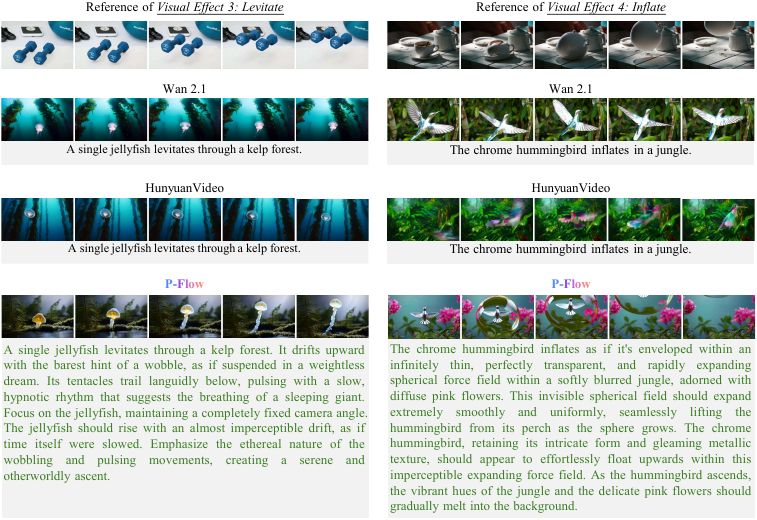}
\vspace{-1.5em}
\caption{Text-to-Video Generation Results.
}
\vspace{-1em}
\label{fig:appendix_results_t2v}
\end{figure*}


\end{document}